\documentclass{article}
\usepackage{spconf,amsmath,graphicx}
\usepackage[pagebackref,breaklinks,colorlinks,bookmarks=false]{hyperref}
\usepackage{amssymb}
\usepackage{booktabs}

\usepackage{multirow}
\usepackage{cite}

\title{HIERARCHICAL INTERACTIVE RECONSTRUCTION NETWORK FOR\\ VIDEO COMPRESSIVE SENSING}

\name{Tong Zhang, Wenxue Cui, Chen Hui, Feng Jiang$^{\ast}$}
\address{School of Computer Science and Technology, Harbin Institute of Technology, Harbin, 150001, China\\State Key Laboratory of Communication Content Cognition, People's Daily Online, Beijing, China 100733 \\ChongQing Research Institute of HIT \leavevmode \\(\{tongzhang, wenxuecui, chenhui\}@stu.hit.edu.cn; fjiang@hit.edu.cn)}

\begin{document}
\maketitle
\begin{abstract}
Deep network-based image and video Compressive Sensing (CS) has attracted increasing attentions in recent years.
However, in the existing deep network-based CS methods, a simple stacked convolutional network is usually adopted, which not only weakens the perception of rich contextual prior knowledge, but also limits the exploration of the correlations between temporal video frames.
In this paper, we propose a novel Hierarchical InTeractive Video CS Reconstruction Network(HIT-VCSNet), which can cooperatively exploit the deep priors in both spatial and temporal domains to improve the reconstruction quality. 
Specifically, in the spatial domain, a novel hierarchical structure is designed, which can hierarchically extract deep features from keyframes and non-keyframes.
In the temporal domain, a novel hierarchical interaction mechanism is proposed, which can cooperatively learn the correlations among different frames in the multi-scale space.
Extensive experiments manifest that the proposed HIT-VCSNet outperforms the existing state-of-the-art video and image CS methods in a large margin.
\end{abstract}
\section{Introduction}
\label{sec:intro}
Compressive Sensing (CS) theory\cite{1580791, 1614066} expounds that if a signal is sparse in a certain domain, it can be recovered from fewer measurements than prescribed by the Nyquist sampling theorem. 
Mathematically, given the initial signal $\mathbf{x} \in \mathbb{R}^N$, the CS measurement $\mathbf{y} \in \mathbb{R}^M$ is obtained by:
\begin{equation}
\mathbf{y} = \mathbf{\Phi} \mathbf{x}
\label{lab:cs}
\end{equation}
where $\mathbf{\Phi} \in \mathbb{R}^{M \times N}$ is the sampling matrix and the sampling ratio can be defined as $\frac{M}{N} \left (M \ll N \right ) $.
CS is widely used in magnetic resonance imaging(MRI)\cite{Lustig2007SparseMT}, snapshot compressive imaging(SCI)\cite{Wang2021MetaSCISA} and image/video coding\cite{Cui2018AnED}.

The core mission of CS is to accurately reconstruct the target signal $\mathbf{x}$ from the compressed measurements $\mathbf{y}$.
Recently, many image CS methods are proposed, which can be roughly categorized as the following two groups: optimization-based methods and deep learning-based methods. 
Specifically, \textbf{1)} optimization-based methods aim to utilize iterative processes to solve a regularized optimization problem:
\begin{equation}
\min _{\mathbf{x}} \frac{1}{2}\|\Phi \mathbf{x}-\mathbf{y}\|_{2}^{2}+\lambda \psi \left ( \mathbf{x} \right ) 
\end{equation}
where the former term $\frac{1}{2}\|\Phi \mathbf{x}-\mathbf{y}\|_{2}^{2}$ denotes the fidelity term and the latter term $\psi \left ( \mathbf{x} \right )$ comes from the prior knowledge, $\lambda$ is the regularization parameter.
The widely used image priors include local smoothing\cite{4060941}, non-local self-similarity\cite{9635679} and sparsity\cite{Xu2018ATW}.
Nevertheless, the high computational complexity limits the practical applications of CS significantly.
\textbf{2)} Deep learning-based methods directly map the measurements to the reconstructed images.
However, these methods generally construct a black box network\cite{2020LAPRAN} which is not interpretable.
In recent years, some deep unfolding networks(DUN)\cite{Cui2022FastHD, Wu2021DenseDU} try to embed deep neural networks into optimization algorithms, such as HQS\cite{Wu2021DenseDU} and ISTA\cite{4959678}.

Recently, CS is successfully applied for the video signal.  
Similar to image CS, video CS reconstruction can also be divided into
optimization-based methods and deep learning-based methods.
For the optimization-based methods, it can be roughly divided into 3D-sparsity reconstruction\cite{2011Accelerated} and motion-compensation reconstruction\cite{2011Residual,2011Video,2017Video}.
The former assumes joint sparsity in the 3D transform domain to recover frames simultaneously.
The latter reconstructs each frame independently by motion compensation.
While the complex computation restricts the practical application seriously.
For the deep learning-based methods, Shi et al.\cite{2021Video} present VCSNet based on CNN and explore both intraframe and interframe correlations. It is noted that the existing image CS methods can be directly used for video frame compression.

 \begin{figure*}
  \centering
    \includegraphics[width=1.0\linewidth]{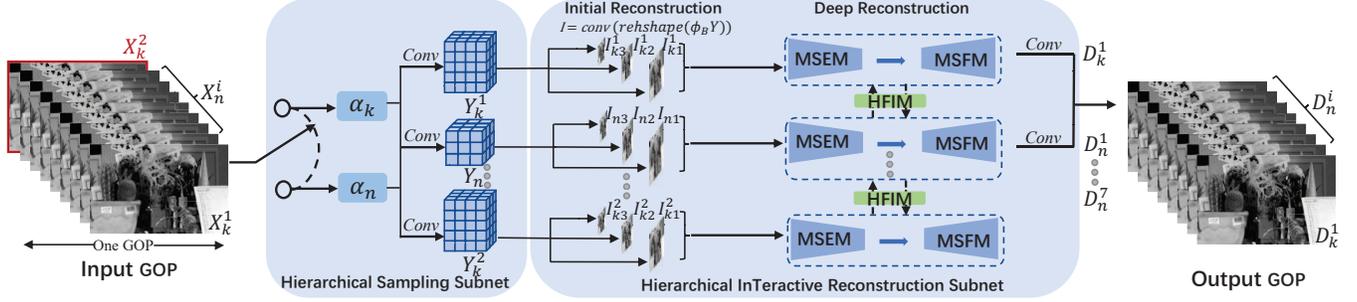}
  \caption{The architecture of the HIT-VCSNet.}
  \label{fig:overview}
\end{figure*}

However, the existing deep learning-based video CS methods still face the following problems:
\textbf{1)} The existing CS networks use simple stacked CNNs, which can not perceive rich spatial contextual prior information effectively.
\textbf{2)} When the motion in the video is fast, it is difficult to capture the temporal correlation efficiently.

To overcome all these drawbacks, in this paper, we propose a novel Hierarchical 
InTeractive Video CS Reconstruction Network(HIT-VCSNet), which can exploit the deep priors in both spatial and temporal domains.
In the spatial domain, we apply a Hierarchical Feature Fusion Module(HFFM) to  hierarchically perceive multi-scale contextual priors.
In the temporal domain, we propose a Hierarchical Feature Interaction Module(HFIM) to automatically interact hierarchical interframe information. 

In summary, our main contributions are as followings:
\textbf{1)} We present an end-to-end Hierarchical InTeractive Video CS Reconstruction Network HIT-VCSNet, which can cooperatively exploit the deep priors in both spatial and temporal domains. 
\textbf{2)} In the spatial domain, a Hierarchical Feature Fusion Module(HFFM) is presented to hierarchically perceive richer contextual priors in the multi-scale space.
\textbf{3)} In the temporal domain, a Hierarchical Feature Interaction Module(HFIM) is developed to automatically learn the interframe correlations in a hierarchical manner.
\textbf{4)} Extensive experiments manifest that the proposed HIT-VCSNet 
outperforms the existing state-of-the-art video and image CS networks in a large margin.

\section{THE PROPOSED HIT-VCSNET}
\label{sec:majhead}
As showed in Fig.\ref{fig:overview}, our HIT-VCSNet composes a hierarchical sampling subnet and a hierarchical interactive reconstruction subnet. 
Given the GOPs as the input of HIT-VCSNet, the sampling subnet outputs the CS measurements of frames.
The initial reconstruction subnet recovers CS measurements into multi-scale initial frames.
The deep reconstruction subnet is composed of a Multi-Scale Extraction Module(MSEM) and a Multi-Scale Fusion Module(MSFM).
Moreover, a HFIM is applied to interact interframe information among keyframes and non-keyframes.
Ultimately, the outputs of MSFM pass a convolutional layer to convert into final frames.

\subsection{Hierarchical Sampling Subnet}
\label{ssec:subhead}
 
\begin{figure}[htb]
\begin{minipage}[b]{1.0\linewidth}
  \centering
  \centerline{\includegraphics[scale=0.7]{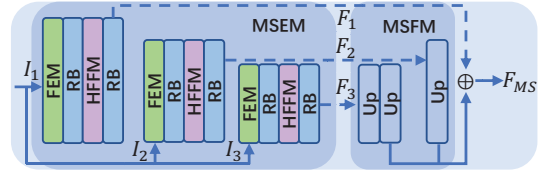}}
  \centerline{(a) Hierarchical Deep Reconstruction Subnet}\medskip
  \label{fig:codec}
\end{minipage}
\begin{minipage}[b]{0.75\linewidth}
  \centering
  \centerline{\includegraphics[scale=0.7]{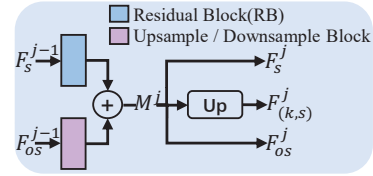}}
  \centerline{(b) Hierarchical Feature Fusion Module}\medskip
  \label{fig:MSFF}
\end{minipage}
\hfill
\begin{minipage}[b]{0.24\linewidth}
  \centering
  \centerline{\includegraphics[scale=0.6]{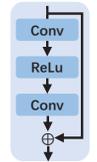}}
  \centerline{(c) RB}\medskip
  \label{fig:RB}
\end{minipage}
\caption{The sub-modules of the deep reconsturction subnet.}
\label{fig:f2}
\end{figure}
As depicted in Fig.\ref{fig:overview}, our sampling subnet has two model settings, the high sampling rate mode for keyframes(i.e.,$\alpha_{k}$) along with the low sampling rate mode for non-keyframes(i.e., $\alpha_{n}$). Therefore, the input frames in a GOP flow into corresponding branches separately.
We divide the input keyframes $\mathbf{X}_{k}$ and non-keyframes $\mathbf{X}_{n}$ into certain numbers of non-overlapping image blocks of size $B \times B$. 
The CS sampling process $\mathbf{y}=\mathbf{\Phi} \mathbf{x}$ can be simulated by convolution as follows:
\begin{equation}
\mathbf{Y}_{k} = \mathbf{S}_{k} * \mathbf{X}_{k}
\end{equation}
\begin{equation}
\mathbf{Y}_{n} = \mathbf{S}_{n} * \mathbf{X}_{n}
\end{equation}
where $\mathbf{S}_{k}$ and $\mathbf{S}_{n}$ has $\alpha_{k} B^2$ and $\alpha_{n} B^2$ filters of size $B \times B$.
In addition, there is no bias term, the stride is B and padding is 0.
The shape of $\mathbf{Y}_{k}$ and $\mathbf{Y}_{n}$ are $h \times w \times \alpha_{k}B^2$ and $h \times w \times \alpha_{n}B^2$.

\subsection{Hierarchical Interactive Reconstruction Subnet}
\label{ssec:subhead}
\textbf{Initial reconstruction:}
\begin{table*}[htbp]
        \small
	\centering
	\caption{Comparison with state-of-the-art video CS methods. The average results of PSNR in dB and SSIM on the first two GOPs of six CIF video sequences with different CS ratios. Best results are in bold.} 
        \setlength{\tabcolsep}{2.9mm}{
	\begin{tabular}{ccccc ccccc}
            \toprule
            Sequence & Ratio & KTSLR\cite{2011Accelerated} & MC/ME\cite{2011Residual} & VideoMH\cite{2011Video} & RRS\cite{2017Video} & VCSNet2\cite{2021Video} & HIT-VCSNet\\
            \midrule 
            Akiyo & \multirow{7}*{0.01} & 29.49/0.8836 & 27.15/0.8191 & 31.16/0.9161 & 25.09/0.7785 & 40.10/0.9834 & \textbf{41.59/0.9842} \\ 
            Coastguard & \multirow{7}*{} & 24.02/0.4902 & 21.65/0.4478 & 24.83/0.5536 & 22.25/0.4304 & 27.93/0.6775 & \textbf{30.75/0.8272} \\ 
            Foreman &\multirow{7}*{} & 24.29/0.7687 & 26.02/0.7504 & 27.84/0.8085 & 20.44/0.6471 & 28.89/0.8283 & \textbf{33.39/0.8949} \\ 
            Mother\_daughter &\multirow{7}*{} & 30.72/0.8592 & 29.29/0.8174 & 32.59/0.8780 & 25.36/0.7142 & 39.80/0.9602 & \textbf{42.18/0.9741} \\
            Paris &\multirow{7}*{} & 20.60/0.5957 & 21.55/0.6616 & 20.64/0.6044 & 17.10/0.4143 & 26.55/0.8668 & \textbf{29.06/0.9263} \\ 
            Silent &\multirow{7}*{} & 25.93/0.7179 & 28.55/0.8327 & 27.22/0.7316 & 22.91/0.5908 & \textbf{34.71/0.9331} & 34.27/0.9300 \\
            Average &\multirow{7}*{} & 25.84/0.7192 & 25.70/0.7215 & 27.38/ 0.7487 & 22.19/0.5959 & 33.00/0.8749 & \textbf{35.21/0.9245} \\
            \midrule 
            Akiyo & \multirow{7}*{0.1} & 33.50/0.9328 & 38.77/0.9657 & 39.48/0.9693 & 41.45/0.9816 & 41.47/0.9815 & \textbf{43.59/0.9904} \\ 
            Coastguard & \multirow{7}*{} & 26.33/0.6222 & 28.09/0.7370 & 28.89/0.7814 & 29.35/0.7843 & 30.24/0.7876 & \textbf{33.13/0.9412} \\ 
            Foreman &\multirow{7}*{} & 28.35/0.8372 & 32.64/0.8803 & 33.94/0.8995 & 35.50/0.9323 & 34.28/0.9222 & \textbf{37.96/0.9786} \\ 
            Mother\_daughter &\multirow{7}*{} & 33.82/0.9064 & 38.56/0.9449 & 38.93/0.9499 & 42.02/0.9719 & 41.43/0.9671 & \textbf{44.07/0.9879} \\
            Paris &\multirow{7}*{} & 23.04/0.7232 & 25.83/0.7778 & 25.58/0.7821 & 24.20/0.8213 & 28.23/0.9030 & \textbf{31.85/0.9819} \\ 
            Silent &\multirow{7}*{} & 29.22/0.8043 & 33.03/0.8841 & 32.69/0.8784 & 35.17/0.9162 & \textbf{36.44/0.9439} & 36.33/0.9428 \\
            Average &\multirow{7}*{} & 29.04/0.8043 & 32.82/0.8650 & 33.25/0.8767 & 34.61/0.9013 & 35.35/0.9176 & \textbf{37.82/0.9705} \\          
            \bottomrule
	\end{tabular}}
\label{table:tab1}
\end{table*}
The initial reconstruction network consists of multiple branches, and each branch corresponds to the frame of each scale.
We set the scale as S. 
For each branch of scale, the initial reconstruction includes three same steps, namely up-sampling, reshape and concatenation:
\begin{equation}
\mathbf{I} = cat(reshape(\mathbf{\Phi}_{B} \mathbf{Y}))
\label{lab:init}
\end{equation}
We utilize a convolutional layer with $B^2$ filters to learn the up-sampling matrix $\mathbf{\Phi}_{B}$, after which a series of vectors of size $1 \times 1 \times B^2$ are generated.
Then a reshape layer is applied to convert each vector to a $B \times B \times 1$ block. 
Ultimately, a concatenation layer is applied to generate the whole frame.

\textbf{Deep reconstruction:}
Our proposed framework jointly extracts features in both spatial and temporal domains.
As depicted in Fig.\ref{fig:f2}(a), MSEM extracts the multi-scale features and MSFM outputs the fused multi-scale feature $\mathbf{F}_{MS}$.
The size of $\mathbf{I}_{1}$, $\mathbf{I}_{2}$ and $\mathbf{I}_{3}$ are 32 × 32, 16 × 16 and 8 × 8.
$\mathbf{F}_{1}$, $\mathbf{F}_{2}$ and $\mathbf{F}_{3}$ correspond to the features of three different scales obtained after MSEM.
Above all, a Feature Extraction Module(FEM) composed of a convolutional layer is applied for each branch to extract the features of the initial frames.
Afterwards, numbers of Residual Blocks(RB) are employed to extract depth features. 
Several up-sampling sub-modules consist MSFM, which is composed of a deconvolution layer with stride 2 as well as two sets of ReLU and convolutional layers.

In the spatial domain, a Hierarchical Feature Fusion Module(HFFM) is implemented in MSEM to perceive richer contextual priors.
As presented in Fig.\ref{fig:f2}(b), for each fusion operation of HFFM at $s-th$ scale, the fused feature $\mathbf{F}_{s}^{j-1}$ from upper level and the feature $\mathbf{F}_{os}^{j-1}$ passed among each scale are merged as $\mathbf{M}^{j}$(j represents the level of RB).
Moreover, up-sampling and down-sampling blocks are implemented due to the scale of the transmitting feature.
The arrangement of the sub-blocks is equal to RB, which involves a skip connection consisting of convolutional layers with stride 1 and a ReLU shown in Fig.\ref{fig:f2}(c). 
Specifically, to obtain the keyframe feature $\mathbf{F}_{(k,s)}^{j}$ for inter-frame fusion, $\mathbf{M}^{j}$ will be passed to an up-sampling block.

In the temporal domain, HFIM is applied to interact hierarchical temporal information from keyframes in current and adjacent GOPs with non-keyframes:
\begin{equation}
\mathbf{F}_{(n,s)}^{(i,j)} = \mathbf{F}_{(k,s)}^{(1,j)} + \mathbf{F}_{(k,s)}^{(2,j)} + RB\left(\mathbf{F}_{(n,s)}^{(i,j-1)}\right)
\end{equation}

\begin{figure}
  \centering
    \centerline{\includegraphics[scale=0.28]{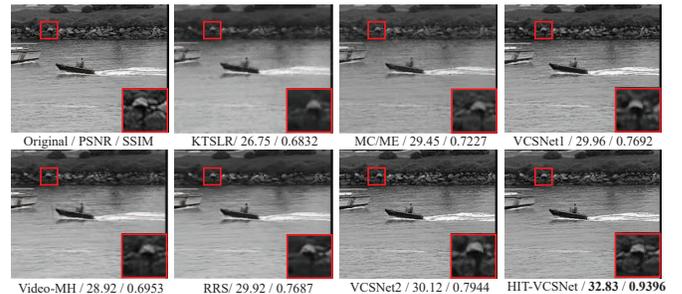}}
    \caption{Visual comparison on the $16^{th}$ frame of selected video sequence $Coastguard$ when the sampling ratio is 0.1.}
    \label{fig:result}
\end{figure}

\subsection{Loss Function}
\label{ssec:subhead}
Given the initial and deep reconstruction of the keyframe and the non-keyframe $\mathbf{I}_{k}$, $\mathbf{D}_{k}$, $\mathbf{I}_{n}$ and $\mathbf{D}_{n}$, the ground-truth keyframe and non-keyframe $\mathbf{X}_{k}$ and $\mathbf{X}_{n}$, we design the end-to-end loss function for HIT-VCSNet as follows:
\begin{equation}
\begin{aligned}
{\scriptsize
\mathcal{L} =\sum_{i=1}^{N}\left(\begin{array}{l}
\left\|\mathbf{I}_{k}^{(i)}-\mathbf{X}_{k}^{(i)}\right\|^{2}_{2}+\left\|\mathbf{D}_{k}^{(i)}-\mathbf{X}_{k}^{(i)}\right\|^{2}_{2}\\
+\left\|\mathbf{I}_{k}^{(i+1)}-\mathbf{X}_{k}^{(i+1)}\right\|^{2}_{2}+\left\|\mathbf{D}_{k}^{(i+1)}-\mathbf{X}_{k}^{(i+1)}\right\|^{2}_{2}\\
+\sum_{k=1}^{K}\left(\left\|\mathbf{I}_{n}^{(i)}-\mathbf{X}_{n}^{(i)}\right\|^{2}_{2}+\left\|\mathbf{D}_{n}^{(i)}-\mathbf{X}_{n}^{(i)}\right\|^{2}_{2}\right)
\end{array}\right)
}
\end{aligned}
\end{equation}
where $K$ refers to the total number of non-keyframes, and $N$ represents the number of GOPs in the training dataset.
\begin{table*}[htbp]
        \small
	\centering
	\caption{ Comparison with deep learning-based image CS methods. The average results of PSNR in dB and SSIM on the first two GOPs of six CIF video sequences($\alpha_{n}=0.1$). Best results are in bold.} 
        \setlength{\tabcolsep}{3mm}{
        \begin{tabular}{cccc ccc}
            \toprule
            Sequence & ISTA-Net$^+$\cite{2017ISTA} & CSNet$^+$\cite{8019428} & SCSNet\cite{8953841} & AMP-Net$^+$\cite{Zhang2021AMPNetDD} & OPINE-Net$^+$\cite{9019857} & HIT-VCSNet \\
            \midrule  
            Akiyo   & 34.83/0.9243     & 35.36/0.9463  &  35.56/0.9567  &  36.57/0.9758  & 36.83/0.9826 & \textbf{43.59/0.9904}  \\ 
            Coastguard  & 27.23/0.6099     & 29.13/0.6946  & 29.33/0.7012  &  30.33/0.7251 & 30.61/0.7357   &  \textbf{33.13/0.9412}     \\ 
            Foreman          & 32.83/0.8795     & 32.38/0.9013  & 32.58/0.9121   &  33.56/0.9341 & 33.87/0.9461   &  \textbf{37.96/0.9786}     \\ 
            Mother\_daughter & 35.54/0.8956     & 37.18/0.9251  & 37.31/0.9310   & 38.36/0.9547 & 38.63/0.9663    & \textbf{44.07/0.9879}     \\ 
            Paris            & 24.07/0.6893     & 24.66/0.7831  &  24.89/0.7931  &  25.83/0.8012 & 26.01/0.8276  & \textbf{31.85/0.9819}      \\ 
            Silent           & 30.23/0.7932     & 31.82/0.8366  &  32.07/0.8451  &  33.06/0.8761 & 33.39/0.8792  & \textbf{36.33/0.9428}     \\ 
            Average          & 30.79/0.8286     & 31.76/0.8478  &  31.96/0.8565  & 32.95/0.8778  & 33.22/0.8896 &\textbf{37.82/0.9705}     \\  
            \bottomrule
        \end{tabular}}
\label{table:tab2}
\end{table*}

\section{EXPERIMENT RESULT}
\label{sec:majhead}

\subsection{Dataset and Implementation Details}
\label{ssec:subhead}
For fair comparison, we use HEVC test video sequences in \cite{2021Video}, which are divided into 128000 groups with data augmentation.
Our HIT-VCSNet is trained end-to-end and the model training is performed on 4 GeForce RTX 3090 GPUs by PyTorch.
We set block size $B=32$, scale $S=3$.  
The GOP size and the batch size are set as 8 and 32 for 100 epochs.
Specifically, we use Adam optimizer with the initial learning rate being $1 \times 10^{-4}$, which is reduced by half after every 30 epochs. 
$\alpha_{k}$ is set to 0.5, and $\alpha_{n}$ is set to 0.01 and 0.1 respectively. As for different sampling rates, the model of non-keyframe is trained independently.
With regards to testing, 6 groups of standard CIF vedio sequences\footnote{Test videos are available at https://media.xiph.org/video/derf/} \textit {Akiyo,
Coastguard, Foreman, Mother\_daughter, Paris, Silent} are applied. 
We utilize standard metrics (PSNR and SSIM\cite{2004Image}) for evaluation. Particularly, we transform color frames into YCbCr space and conduct operation merely for Y channel(i.e., luminance).

\begin{table}[htbp]
        \scriptsize
	\centering
	\caption{ The average results of PSNR on the first two GOPs compared with image CS methods ($\alpha_n$ = 0.01).} 
        \setlength{\tabcolsep=0.14cm}{
        \begin{tabular}{cccc ccc}
            \toprule
            Sequence  & SCSNet\cite{8953841} & AMP-Net$^+$\cite{Zhang2021AMPNetDD} & OPINE-Net$^+$\cite{9019857} & HIT-VCSNet \\
            \midrule  
            Average  & 26.38 & 27.75  & 28.83  &\textbf{35.21}     \\         
            \bottomrule
        \end{tabular}}
\label{table:tab3}
\end{table}
\begin{table}[htbp]
        \scriptsize
	\centering
        \caption{ The average results of PSNR on the non-keyframes of the first two GOPs compared with image CS methods.}
        \setlength{\tabcolsep=0.23cm}{
	\begin{tabular}{cccc ccccc}
            \toprule
            Ratio &  ISTA-Net$^+$\cite{2017ISTA} & CSNet$^+$\cite{8019428} & VCSNet2\cite{2021Video} & HIT-VCSNet\\
            \midrule            
            0.01  & 20.14 & 24.14 & 31.95 & \textbf{33.95} \\
            \midrule
            0.1  & 29.34 & 30.53 & 34.63 & \textbf{36.56} \\          
            \bottomrule
	\end{tabular}}
\label{table:tab4}
\end{table}

\subsection{Comparisons with State-of-the-Art Methods}
\label{ssec:subhead} 
    We evaluate our HIT-VCSNet  with state-of-the-art video CS methods, including KTSLR\cite{2011Accelerated}, MC/ME\cite{2011Residual},
Video-MH\cite{2011Video},  RRS\cite{2017Video} and VCSNet2\cite{2021Video}.
Moreover, we compare five deep learning-based image CS methods with HIT-VCSNet, namely ISTA-Net\cite{2017ISTA}, CSNet\cite{8019428}, SCSNet\cite{8953841}, AMPNet\cite{Zhang2021AMPNetDD} and OPINE-Net\cite{9019857}.
As reported in Table \ref{table:tab1}-Table \ref{table:tab4} and Fig.\ref{fig:result}, one can see that our HIT-VCSNet outperforms all the other methods.
In particular, due to the tiny range of motion, the reconstruction effect of HIT-VCSNet is worse than VCSNet2 on the \textit{silent} sequence, while our model owns predominant performance for large scale motion.
Moreover, the corresponding number of network parameters and time consumption of HIT-VCSNet are 4-5 times that of VCSNet2.

\begin{table}[htbp]
  \small
  \centering
  \caption{Ablation study of HFIM and HFFM modules.}
  \setlength{\tabcolsep}{3.5mm}{
  \begin{tabular}{cccc}
    \toprule
    Ratio & w/o HFIM & w/o HFFM & HIT-VCSNet \\
    \midrule
    0.01 & 33.96/0.8917 & 32.99/0.8792 & \textbf{35.21/0.9240} \\
    0.1 & 36.53/0.9372 & 35.72/0.9267 & \textbf{37.82/0.9705} \\
    \bottomrule
  \end{tabular}
  }
  \label{table:ablation}
\end{table}
\subsection{Ablation Study}
\label{ssec:subhead} 
We retrain our model without HFIM and HFFM respectively, represented as “w/o HFIM” and “w/o HFFM”.
We evaluate the average PSNR and SSIM from the models for the first two GOPs of the 6 test video sequences. 
As shown in Table \ref{table:ablation}, HIT-VCSNet leads to a boost of 1.25dB and 2.22dB on PSNR at the sampling
ratio of 0.01 compared with “w/o HFIM” and “w/o HFFM”, which reflects the effectiveness of exploiting the deep priors in both spatial and temporal domains.

\section{CONCLUSION}
\label{sec:majhead}
In this paper, we propose a novel Hierarchical InTeractive Video CS Reconstruction Network HIT-VCSNet, which can cooperatively exploit the deep priors in both spatial and temporal domains.
Moreover, the hierarchical structure enables the proposed framework not only to hierarchically exploit richer contextual priors, but also to capture the interframe correlations more efficiently in the multi-scale space.
Extensive experiments manifest that the proposed HIT-VCSNet outperforms the existing state-of-the-art video and image CS methods in a large margin.

\subsection*{Acknowledgments}
This work is founded by National Natural Science Foundation of China (No.62076080), Natural Science Foundation of ChongQing CSTB2022NSCQ-MSX0922 and the Postdoctoral Science Foundation of Heilongjiang Province of China (LBH-Z22175).


\vfill\pagebreak


\bibliographystyle{plain}
\bibliography{ref}

\end{document}